\documentclass[conference]{IEEEtran}
\usepackage{cite}
\usepackage{enumitem}
\usepackage{amsmath,amssymb,amsfonts}
\usepackage{algorithmic}
\usepackage{graphicx}
\usepackage{textcomp}
\usepackage{xcolor}
\def\BibTeX{{\rm B\kern-.05em{\sc i\kern-.025em b}\kern-.08em
    T\kern-.1667em\lower.7ex\hbox{E}\kern-.125emX}}
\begin{document}

\title{A Hybrid Architecture for Benign-Malignant Classification of Mammography ROIs} 

\author{
\centering
\begin{tabular}{c@{\hspace{2cm}}c}

\begin{tabular}{c}
\textbf{Mohammed Asad}\\
Dept. of Electronics and\\
Communication Engineering\\
Delhi Technological University\\
Delhi, India\\
asadmohd9411@gmail.com
\end{tabular}
&
\begin{tabular}{c}
\textbf{Mohit Bajpai}\\
Dept. of Electronics and\\
Communication Engineering\\
Delhi Technological University\\
Delhi, India\\
mohitbajpai998@gmail.com
\end{tabular}

\\[1.5em]

\begin{tabular}{c}
\textbf{Sudhir Singh}\\
Dept. of Information Technology\\
IGDTUW\\
Delhi, India\\
sudhirsingh@igdtuw.ac.in
\end{tabular}
&
\begin{tabular}{c}
\textbf{Rahul Katarya}\\
Dept. of Computer Science\\
and Engineering\\
Delhi Technological University\\
Delhi, India\\
rahulkatarya@dtu.ac.in
\end{tabular}

\end{tabular}
}

\maketitle

\begin{abstract}
Accurate characterization of suspicious breast lesions in mammography is important for early diagnosis and treatment planning. While Convolutional Neural Networks (CNNs) are effective at extracting local visual patterns, they are less suited to modeling long-range dependencies. Vision Transformers (ViTs) address this limitation through self-attention, but their quadratic computational cost can be prohibitive. This paper presents a hybrid architecture that combines EfficientNetV2-M for local feature extraction with Vision Mamba, a State Space Model (SSM), for efficient global context modeling. The proposed model performs binary classification of abnormality-centered mammography regions of interest (ROIs) from the CBIS-DDSM dataset into benign and malignant classes. By combining a strong CNN backbone with a linear-complexity sequence model, the approach achieves strong lesion-level classification performance in an ROI-based setting.
\end{abstract}


\section{Introduction}
Breast cancer remains a leading cause of cancer-related mortality among women globally, making early detection paramount for improving survival rates. Mammography is the gold standard for screening due to its accessibility and cost-effectiveness. However, its interpretation is challenging; high breast density can obscure lesions, and subtle morphological signs of early-stage cancers can be missed, leading to inter-observer variability among radiologists \cite{b1}.

Computer-Aided Diagnosis (CADx) systems have evolved significantly to address these challenges. Early systems relied on handcrafted features \cite{b2}, but deep learning, particularly Convolutional Neural Networks (CNNs) like EfficientNet \cite{b3}, revolutionized the field by automating feature extraction and achieving performance approaching human experts. However, the localized receptive fields of CNNs limit their ability to model long-range spatial dependencies crucial for identifying abnormalities like architectural distortions.

Vision Transformers (ViTs) were introduced to capture global context using self-attention mechanisms \cite{b4}. However, their quadratic computational complexity ($O(N^2)$) makes them inefficient for high-resolution mammography images. This has spurred the development of State Space Models (SSMs) like Mamba \cite{b5} and its vision-centric variant, Vision Mamba (ViM) \cite{b6}, which model long sequences with linear complexity ($O(N)$), offering Transformer-like global reach without prohibitive computational costs.

\noindent This paper presents:
\begin{itemize}[leftmargin=1.5em,nosep,topsep=0pt,parsep=0pt]
    \item A hybrid architecture combining EfficientNetV2-M for local feature extraction with Vision Mamba for efficient global context modeling.
    \item An ROI-based benign--malignant classification framework for abnormality-centered mammography crops from CBIS-DDSM.
    \item A comparative evaluation against CNN and Transformer baselines under a common preprocessing pipeline.
    \item An ablation study assessing the contribution of the CNN and SSM components.
\end{itemize}

\section{Dataset and Preprocessing}

\subsection{CBIS-DDSM Dataset}
The Curated Breast Imaging Subset of the Digital Database for Screening Mammography (CBIS-DDSM) is a widely used public dataset for computer-aided diagnosis of breast cancer \cite{b7}. It includes mammography studies with pathology information and expert annotations. In this work, we use abnormality-centered cropped regions of interest (ROIs) associated with annotated lesions rather than full mammograms.

We formulate the task as binary classification, grouping malignant lesions into a single malignant class (label = 1) and benign lesions into a single benign class (label = 0). The resulting ROI-based setting focuses on lesion-level classification rather than whole-mammogram screening. Key challenges include:
\begin{itemize}
    \item Variation in lesion appearance, scale, and texture across benign and malignant cases.
    \item Class imbalance, addressed through stratified splitting and weighted loss.
\end{itemize}

CBIS-DDSM is a relevant benchmark for evaluating the proposed EfficientNetV2-M--Vision Mamba architecture in an ROI-based mammography classification setting.

\subsection{Data Preparation Pipeline}
The raw CBIS-DDSM data was processed through a reproducible pipeline to produce tensors for training and evaluation:
\begin{enumerate}[label=\textbf{\alph*)}, leftmargin=1.5em]
    \item \textbf{Archive extraction and file discovery:} Archives were unpacked into a consistent folder structure. A recursive scan located all JPEG images, logging counts and flagging irregularities.
    \item \textbf{Identifier extraction and path matching:} Stable identifiers from metadata CSV files were matched to JPEG images. Unmatched entries were excluded to prevent incomplete samples.
    \item \textbf{Label encoding and sanity checks:} Diagnostic labels were mapped to binary codes: 1 for malignant, 0 for benign. Class distribution was inspected to flag inconsistencies.
    \item \textbf{Deterministic image loading and channel handling:} Images were read using a fixed PIL loader and converted to three-channel RGB by duplicating the grayscale channel to meet pretrained backbone requirements.
    \item \textbf{Transformations and normalization:} Images were resized to 224 $\times$ 224 pixels using bicubic interpolation, converted to tensors, and normalized with ImageNet mean and standard deviation. Light augmentations (horizontal flips, $\pm$10$^\circ$ rotation) were applied to the training set only.
    \item \textbf{Train/validation split and loaders:} A patient-level stratified split (70\%/15\%/15\%) maintained class balance and avoided subject overlap. Custom Dataset and DataLoader classes optimized batching and I/O.
    \item \textbf{Checkpointing and persistence:} Split definitions and model checkpoints were stored for reproducibility.
\end{enumerate}

\subsection{Data Augmentation}
To mitigate overfitting, lightweight, on-the-fly augmentations were applied during training:
\begin{itemize}
    \item \textbf{Horizontal flip:} Random left-right mirroring (\texttt{RandomHorizontalFlip}).
    \item \textbf{Small-angle rotation:} Random rotations of $\pm$10$^\circ$ (\texttt{RandomRotation}).
    \item \textbf{Split-specific application:} Augmentations applied only to the training set; validation and test sets remained unchanged.
\end{itemize}
These transformations encourage focus on diagnostically meaningful features while preserving core lesion characteristics \cite{b9}.

\section{Proposed Hybrid Architecture}
The proposed framework integrates a high-capacity Convolutional Neural Network (CNN) backbone for local feature extraction with a State Space Model (SSM) for efficient global context modeling, mirroring the clinical workflow where radiologists identify local patterns before assessing broader anatomical context.

\subsection{EfficientNetV2 for Local Feature Extraction}
EfficientNetV2 is chosen as the convolutional backbone for its strong trade-off between accuracy, parameter efficiency, and training speed \cite{b10}. Its Fused-MBConv block replaces separate depthwise and expansion convolutions with a single 3$\times$3 convolution in early layers, improving throughput on modern accelerators. EfficientNet architectures have shown strong mammography performance, with prior work achieving an AUC of 0.848 on CBIS-DDSM \cite{b3}. We use EfficientNetV2-M, removing its classification head to output a multi-scale feature representation from the final convolutional stage.

\subsection{Vision Mamba for Global Context Modeling}
\begin{figure}[htbp]
    \centering
    \includegraphics[width=\columnwidth]{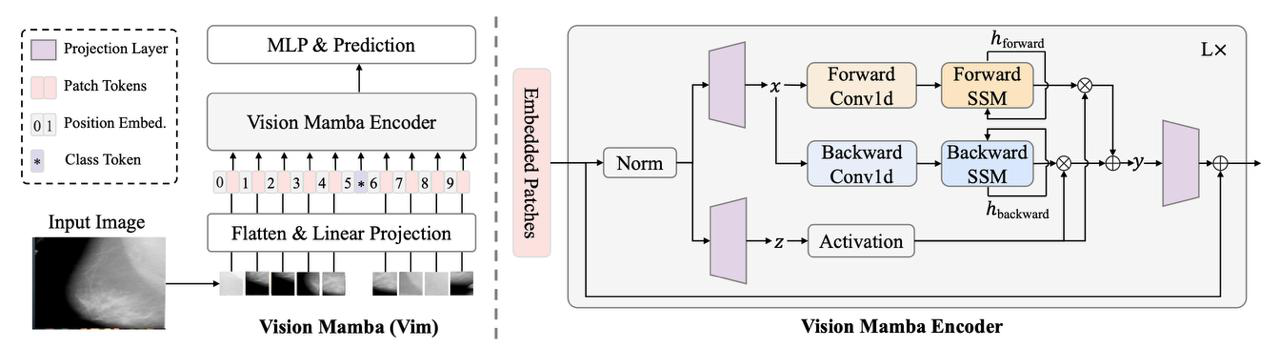}
    \caption{Vision Mamba architecture, illustrating bidirectional state space modeling for efficient global context capture.}
    \label{fig:vim}
\end{figure}

Vision Mamba (ViM) \cite{b6} is a vision-focused State Space Model that overcomes the quadratic complexity of Transformer self-attention. Operating in linear time and space ($O(L)$), ViM is ideal for high-resolution medical images. It tokenizes image features into patch sequences and processes them through bidirectional Mamba blocks, scanning forward and backward to capture global dependencies efficiently, as shown in Fig.~\ref{fig:vim}. Prior work, such as MedMamba \cite{b11}, demonstrates the efficacy of hybrid CNN--SSM pipelines in medical imaging.

\subsection{Architectural Fusion}
\begin{figure*}[!t]
    \centering
    \includegraphics[width=\textwidth]{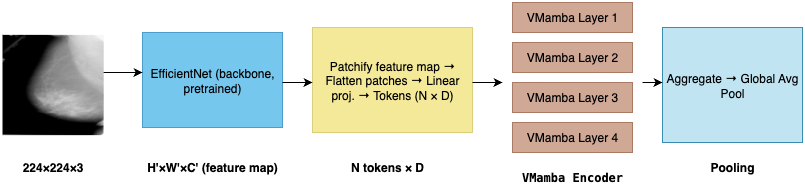}
    \caption{Proposed hybrid architecture combining EfficientNetV2-M for local feature extraction and Vision Mamba for global context modeling.}
    \label{fig:hybrid}
\end{figure*}

The EfficientNetV2 and Vision Mamba components form a sequential pipeline for binary classification, as depicted in Fig.~\ref{fig:hybrid}:
\begin{enumerate}
    \item \textbf{Feature Map Extraction:} The input ROI image (224$\times$224 pixels, three channels, normalized) is passed through EfficientNetV2-M (pretrained on ImageNet), with the classification head removed. The final convolutional layer outputs a feature map of size 12$\times$12$\times$1280.
    \item \textbf{Feature Map Patchification:} The feature map is divided into non-overlapping patches, each flattened into a vector (token), forming a 1D sequence \cite{b4}.
    \item \textbf{Positional Encoding:} Learnable positional embeddings preserve spatial context lost during flattening.
    \item \textbf{Global Context Modeling:} The token sequence is processed by bidirectional Vision Mamba blocks for long-range dependency modeling.
    \item \textbf{Classification Head:} The output sequence is aggregated via Global Average Pooling (GAP) \cite{b12} to produce a feature vector, passed through a fully connected layer with sigmoid activation to yield malignancy probability.
\end{enumerate}

This architecture leverages EfficientNetV2-M for precise local feature extraction and Vision Mamba for efficient global reasoning in ROI-based mammography classification.

\section{Experimental Setup}

\subsection{Training Strategy}
A transfer learning workflow initialized the EfficientNetV2-M backbone with ImageNet-pretrained weights, providing robust low- and mid-level visual features. Vision Mamba layers and the classification head were randomly initialized for adaptation to mammography. Training used 50 epochs, batch size 16, on an NVIDIA A100 GPU, with early stopping based on validation AUC.

Training proceeded in two stages:
\begin{itemize}
    \item \textbf{Feature extraction phase:} EfficientNetV2-M layers were frozen, training only Vision Mamba layers and the classification head to learn task-specific decision boundaries and global context.
    \item \textbf{Fine-tuning phase:} All layers were unfrozen, with a reduced learning rate for the backbone, enabling adaptation to domain-specific patterns.
\end{itemize}
This strategy accelerates convergence, reduces overfitting, and combines prior visual knowledge with task-specific features.

\subsection{Optimizer and Loss Function}
\begin{enumerate}[label=\textbf{\alph*)}, leftmargin=1.5em]
    \item \textbf{Optimizer:} AdamW was chosen for decoupling weight decay from gradient updates, improving stability and generalization \cite{b13}.
    \item \textbf{Learning-rate schedule:} Initial learning rates were $3\times10^{-4}$ for new layers and $3\times10^{-5}$ for the backbone during fine-tuning, with cosine annealing and warm restarts \cite{b14}.
    \item \textbf{Loss function:} Binary cross-entropy with class weights inversely proportional to class frequencies mitigated imbalance \cite{b19}.
\end{enumerate}

\subsection{Evaluation Metrics}
Performance was evaluated on a held-out test set using:
\begin{itemize}
    \item \textbf{Area under the ROC curve (AUC-ROC):} Measures ability to distinguish malignant from benign cases, independent of threshold \cite{b20}.
    \item \textbf{Sensitivity (Recall):} Proportion of malignant cases correctly identified, critical to minimize missed cancers \cite{b21}.
    \item \textbf{Specificity:} Proportion of benign cases correctly identified, reducing unnecessary procedures \cite{b21}.
    \item \textbf{F1-score and Accuracy:} Balance precision and recall, addressing class imbalance.
\end{itemize}

\section{Results and Discussion}

\subsection{Performance Comparison}
The proposed model was benchmarked against several baseline architectures on the CBIS-DDSM test set. Table~\ref{tab:cbis_ddsm_comparison_alt} reports parameter count, AUC, accuracy, sensitivity, specificity, and F1-score. Baselines marked with an asterisk were re-implemented under the same preprocessing pipeline.

\begin{table}[!t]
\caption{Performance Comparison on the CBIS-DDSM Test Set}
\centering
\scriptsize
\setlength{\tabcolsep}{3pt}
\resizebox{\columnwidth}{!}{%
\begin{tabular}{lcccccc}
\hline
\textbf{Model} & \textbf{Params (M)} & \textbf{AUC} & \textbf{Acc. (\%)} & \textbf{Sens.} & \textbf{Spec.} & \textbf{F1} \\
\hline
VGG-16 \cite{b22}* & 138.0 & 0.795 & 87.2 & 0.78 & 0.89 & 0.80 \\
ResNet-50 \cite{b23}* & 23.5 & 0.812 & 89.1 & 0.80 & 0.90 & 0.82 \\
Inception-v3 \cite{b24}* & 23.8 & 0.819 & 89.8 & 0.81 & 0.91 & 0.83 \\
DenseNet-121 \cite{b25}* & 7.0 & 0.825 & 90.3 & 0.82 & 0.92 & 0.84 \\
ViT-B/16 \cite{b4}* & 86.0 & 0.831 & 90.8 & 0.83 & 0.92 & 0.85 \\
Two-View EfficientNet-B0 \cite{b3} & 5.3 & 0.848 & 92.1 & 0.85 & 0.93 & 0.87 \\
Swin-T \cite{b26}* & 28.0 & 0.842 & 91.5 & 0.84 & 0.93 & 0.86 \\
EfficientNet-B3 \cite{b27}* & 12.0 & 0.855 & 92.8 & 0.86 & 0.94 & 0.88 \\
CMT-Small \cite{b28}* & 25.2 & 0.861 & 93.3 & 0.87 & 0.94 & 0.89 \\
\textbf{Proposed: EffNetV2-M + ViM} & \textbf{32.4} & \textbf{0.875} & \textbf{94.2} & \textbf{0.89} & \textbf{0.95} & \textbf{0.90} \\
\hline
\multicolumn{7}{l}{\footnotesize * Re-implemented on CBIS-DDSM with the same preprocessing pipeline.}
\end{tabular}%
}
\label{tab:cbis_ddsm_comparison_alt}
\end{table}

On our ROI-based CBIS-DDSM test split, the proposed EfficientNetV2-M + Vision Mamba model achieved an AUC of 0.875, accuracy of 94.2\%, sensitivity of 0.89, specificity of 0.95, and F1-score of 0.90. Among the evaluated baselines, it produced the strongest overall results. The improvement over EfficientNet-B3 and Swin-T suggests that combining local CNN features with efficient long-range sequence modeling is beneficial in this lesion-level classification setting. Comparisons with two-view methods such as \cite{b3} should be interpreted with caution, as they use a different input protocol.

\subsection{Ablation Study}
To quantify the hybrid architecture’s benefits, we evaluated its components independently. Table~\ref{tab:ablation} compares the proposed model against standalone EfficientNetV2-M and Vision Mamba.

\begin{table}[!t]
\caption{Ablation Study on the CBIS-DDSM Test Set}
\centering
\begin{tabular}{lccccc}
\hline
\textbf{Model} & \textbf{AUC} & \textbf{Acc. (\%)} & \textbf{Sens.} & \textbf{Spec.} & \textbf{F1} \\
\hline
EfficientNetV2-M & 0.850 & 92.5 & 0.85 & 0.93 & 0.87 \\
Vision Mamba & 0.820 & 89.0 & 0.80 & 0.90 & 0.82 \\
\textbf{EffNetV2-M + ViM} & \textbf{0.875} & \textbf{94.2} & \textbf{0.89} & \textbf{0.95} & \textbf{0.90} \\
\hline
\end{tabular}
\label{tab:ablation}
\end{table}

The hybrid model improves AUC by 2.5\% over EfficientNetV2-M alone, confirming the value of Vision Mamba’s global context modeling.

\section{Limitations}
This study is conducted in an ROI-based setting using abnormality-centered cropped mammography images rather than full mammograms, which makes the task different from screening-level breast cancer detection. In addition, the present study uses single-view lesion crops, whereas some prior methods exploit two-view information (CC and MLO). Future work will examine full-image and multi-view modeling.

\section{Conclusion}
This paper presents a hybrid architecture that combines EfficientNetV2-M for local feature extraction with Vision Mamba for efficient long-range dependency modeling in ROI-based benign--malignant mammography classification on CBIS-DDSM. The proposed model achieved an AUC of 0.875, accuracy of 94.2\%, sensitivity of 0.89, specificity of 0.95, and F1-score of 0.90, yielding the strongest results among the evaluated baselines in our experimental setup. These findings suggest that integrating CNN-based local representation learning with Mamba-based global context modeling is promising for lesion-level mammography classification. Future work will investigate multi-view modeling, external validation, and more comprehensive reproducibility analysis.

\section*{Acknowledgment}
The authors thank the maintainers of the CBIS-DDSM dataset and the open-source deep learning community for their contributions and resources.


\end{document}